\documentclass[10pt,twocolumn,letterpaper]{article}

\usepackage{wacv}
\usepackage{times}
\usepackage{epsfig}
\usepackage{graphicx}
\usepackage{amsmath}
\usepackage{amssymb}
\usepackage[normalem]{ulem}
\usepackage{gensymb}

\usepackage{subcaption}
\usepackage{flushend}

\def\smean#1{\overline{#1}}
\def\svar#1{\overline{\overline{#1}}}

\newcommand{\add}[1]{{\color{black} #1}}

\wacvfinalcopy 

\def\wacvPaperID{206} 

\ifwacvfinal\fi
\begin{document}

\title{Domain-Specific Human-Inspired\\ Binarized Statistical Image Features for Iris Recognition\thanks{Paper accepted for WACV 2019, Hawaii, USA}}

\author{Adam Czajka \hspace{1cm} Daniel Moreira \hspace{1cm} Kevin W. Bowyer \hspace{1cm} Patrick J. Flynn \\
Department of Computer Science and Engineering, University of Notre Dame\\
{\tt\small \{aczajka,dhenriq1,kwb,flynn\}@nd.edu}
}

\maketitle
\ifwacvfinal\thispagestyle{empty}\fi

\begin{abstract}
Binarized statistical image features (BSIF) have been successfully used for texture analysis in many computer vision tasks, including iris recognition and biometric presentation attack detection. One important point is that all applications of BSIF in iris recognition have used the original BSIF filters, which were trained on image patches extracted from natural images. This paper tests the question of whether domain-specific BSIF can give better performance than the default BSIF. The second important point is in the selection of image patches to use in training for BSIF. Can image patches derived from eye-tracking experiments, in which humans perform an iris recognition task, give better performance than random patches? Our results say that (1) domain-specific BSIF features can out-perform the default BSIF features, and (2) selecting image patches in a task-specific manner guided by human performance can out-perform selecting random patches. These results are important because BSIF is often regarded as a generic texture tool that does not need any domain adaptation, and human-task-guided selection of patches for training has never (to our knowledge) been done. This paper follows the reproducible research requirements, and the new iris-domain-specific BSIF filters, the patches used in filter training, the database used in testing and the source codes of the designed iris recognition method are made available along with this paper to facilitate applications of this concept.
\end{abstract}

\section{Introduction}
\label{sec:intro}

Binarized Statistical Image Features (BSIF) have been shown to be effective in iris recognition \cite{Raja_IWBF_2014,rathgeb_2016} as well as in iris presentation attack detection \cite{Doyle_ACCESS_2015,Ghiani_BTAS_2013,Raghavendra_TIFS_2015}. All these approaches applied the standard BSIF filters provided with the original paper \cite{Kannala_ICPR_2012}.
These filters originate from a different domain than iris recognition. 
Using these filters assumes that filter kernels developed for a small set of natural images can serve as universal feature extractors independently of the application. However, we hypothesize that a) preparation of domain-specific filters, employing image patches sampled from a new domain for filter training, allows to extract features that are more discriminative for the specific domain, and b) observing how humans perform iris recognition task helps in building more powerful feature extractors. Specifically, this paper answers the following questions:

\begin{enumerate}
    \item[Q1.] Does the adaptation of BSIF filters to an iris recognition domain allow to extract more discriminative iris image features than standard BSIF filters?
    \item[Q2.] Does a careful selection of iris image patches, based on regions used by humans performing iris recognition task, allow to increase the iris recognition performance when compared to using filters trained on randomly selected iris images patches?
\end{enumerate}

To calculate new iris-domain specific filters, feedback from human subjects was used to select the most salient training samples.
Each of 86 subjects, who participated in the experiments, was presented with 10 iris image pairs (randomly selected out of 160 different pairs) and their task was to:
(a) decide if each pair of images presents the same iris; during this step, the gaze of subjects was automatically collected by the means of an eye tracker device, and
(b) annotate regions of the images supporting their decision.
As a consequence, we compiled two sets of iris image patches deemed to be salient for matching iris images: one coming from the gaze of subjects, and the other coming from the provided annotations. For comparison purposes, the third set of new iris-domain specific filters was designed based on patches selected randomly from iris images, \ie without involving humans.

\begin{figure*}[!htb]
    \centering
        \includegraphics[width=\textwidth]{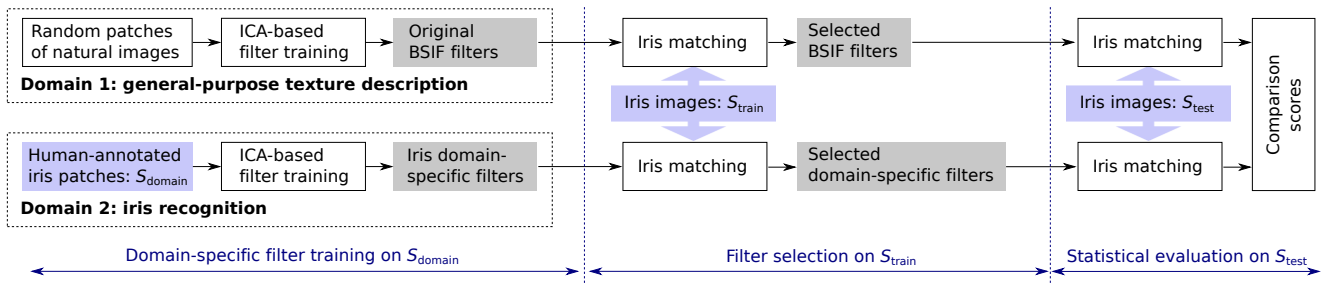}
    \caption{General pipeline of domains-specific filter re-training, selection and evaluation.}
    \label{fig:generalPipeline}
\end{figure*}

To verify the hypothesis that domain-specific filters can outperform the standard BSIF filters, we applied a three-stage procedure incorporating specially designed subject-disjoint data sets, and statistical testing to verify if the observed differences among distinct approaches are statistically significant, as depicted in Fig. \ref{fig:generalPipeline}:

\begin{enumerate}
\item The first stage defines the application domain. The standard BSIF filters were trained on patches of natural images. Our filters are trained on human-selected iris image patches extracted from the $\mathcal{S}_{\mbox{\scriptsize domain}}$ set of images. To maximize the heterogeneity of the sample iris images, $\mathcal{S}_{\mbox{\scriptsize domain}}$ includes image pairs representing different tails of the comparison score distributions (``easy'' samples and ``hard'' samples), iris images of twins, samples with large difference in pupil dilation, and even post-mortem iris images. Four different sensors (LG 4000, AD 100, IriShield MK 2120U, and laboratory prototype based on TheImagingSource DMK camera) were used in total to collect images included into $\mathcal{S}_{\mbox{\scriptsize domain}}$ to prevent new filters to be sensor-specific, rather than iris-domain-specific.

\item In the second stage, all the required hyper-parameters are selected for the classification methods using the standard and our newly-designed filters. We use a separate set $\mathcal{S}_{\mbox{\scriptsize train}}$ of iris images, subject-disjoint with the $\mathcal{S}_{\mbox{\scriptsize domain}}$ set, to select filter sets delivering the most discriminating features in iris domain.

\item The third stage encompasses the statistical evaluation and comparison among the methods, and the third $\mathcal{S}_{\mbox{\scriptsize test}}$ set, subject-disjoint with the $\mathcal{S}_{\mbox{\scriptsize domain}}$ and $\mathcal{S}_{\mbox{\scriptsize train}}$ sets, is used to make this comparison sound from a statistical point of view.
\end{enumerate}

\noindent The above rigorous procedure, incorporating subject-disjoint data used in each stage, allows to follow a typical scenario of how standard BSIF filters are currently applied in computer vision problems, and hence minimizes the risk of biased evaluations.

There are three main {\bf contributions of this paper}:
\begin{enumerate}
    \item Experiments and comprehensive evaluation showing that a) domain-specific filters extract features that are more powerful in the domain at hand than standard BSIF filters, and b) using human feedback in the filter design process increase chances to obtain further increase in discrimination power of the domain-specific filters.
    
    \item Domain-specific (new) filters ready to be applied in the standard BSIF pipeline for various tasks related to iris recognition, along with iris image patches used in filter re-training and testing database\footnote{The source codes, iris image patches and new retrained BSIF filters are available at \url{https://github.com/CVRL/domain-specific-BSIF-for-iris-recognition}. Please follow the instructions at \url{https://cvrl.nd.edu/projects/data/} to get a copy of the test database.\label{footnote:filters}}.
    
    \item Source codes of the iris recognition method using domain-specific BSIF filters that offers better accuracy than other open-source BSIF-based and Gabor-based iris recognition methods\textsuperscript{\ref{footnote:filters}}.
\end{enumerate}

The main purpose of this paper is to show, to our knowledge for the first time, that we may benefit from human feedback when building feature extractors for iris recognition. Next section defines the context of this paper and presents iris recognition methods based on BSIF. Sec.~\ref{sec:rw} summarizes the related work. In Sec.~\ref{sec:training}, we present the experimental setup designed to build $\mathcal{S}_{\mbox{\scriptsize domain}}$ set and to train new filters, while Sec.~\ref{sec:selection} provides the filter selection methodology. Sec.~\ref{sec:results} presents statistical analysis of the results obtained for standard and new filters, as well as the comparison with other open-source iris recognition methods (one BSIF-based and one incorporating Gabor filtering) on the same test set. Finally, in Sec.~\ref{sec:conclusions}, we conclude the paper and elaborate on future work.

\section{Domain Definition: Iris Recognition}
\label{sec:irisRecognition}

Iris textures observed in near-infrared light are subjectively rather different from textures in natural images.
This paper focuses on iris recognition as a specific example domain for which we design new domain-specific BSIF filters. 

\begin{figure*}[!htb]
    \centering
        \includegraphics[width=\textwidth]{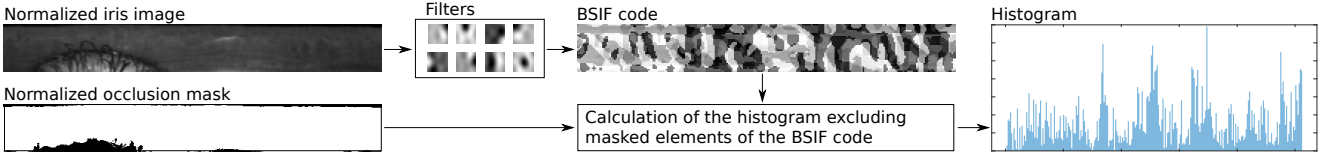}
    \caption{Iris recognition pipeline based on BSIF histograms calculated for non-occluded iris areas. In this example, $n=8$ filters of size $l = 17$ to generate the BSIF code were used. {\bf This histogram-based iris matching approach is added to this paper for comparison purposes only}.}
    \label{fig:iris_rec_pipeline_hist}
\end{figure*}

There are various ways of how the BSIF pipeline can be adapted for iris recognition. A straightforward way, proposed earlier in the literature \cite{USIT2}, is to use the histogram, either normalized or raw, of the composed filtering results as an iris image template, see Fig. \ref{fig:iris_rec_pipeline_hist}; this is termed the ``BSIF code''~\cite{Kannala_ICPR_2012}. In this approach, the unwrapped iris image is filtered by a set of $n$ filters of size $l\times l$ pixels. Each filter response is then binarized (with a threshold at zero), and the $n$ resulting bits for each pixel are translated into the $n$-bit grey-scale value, and finally all gray-scale values calculated for the entire image are represented as a histogram with $2^n$ bins. In our implementation of this approach, we use an occlusion mask to exclude regions of the BSIF code that do not correspond to the iris texture, and we use only valid portions of the convolution results when calculating the histograms. Finally, the comparison score between iris templates is the $\chi^2$ distance between raw template histogram $h^{(t)}$ and raw probe histogram $h^{(p)}$:

\begin{equation}
z_{\mbox{\scriptsize HistRaw}} = \frac{1}{2}\sum_{i=0}^{2^n-1}\frac{\Big(h_i^{(t)}-h_i^{(p)}\Big)^2}{h_i^{(t)}+h_i^{(p)}+\epsilon},
\label{zh}
\end{equation}

\noindent where $h_i$ is the $i$-th bin of the histogram $h$, and $\epsilon$ is a small number that prevents from division by zero when $h_i^{(t)}=h_i^{(p)}=0$. As in the original BSIF pipeline, we have additionally considered normalized histograms $\hat{h}$ to calculate an alternative comparison score:

\begin{equation}
z_{\mbox{\scriptsize HistNormalized}} = \frac{1}{2}\sum_{i=0}^{2^n-1}\frac{\Big(\hat{h}_i^{(t)}-\hat{h}_i^{(p)}\Big)^2}{\hat{h}_i^{(t)}+\hat{h}_i^{(p)}+\epsilon},
\label{zhn}
\end{equation}

\noindent where

$$
\hat{h}_i = \frac{h_i}{\sum_{i=0}^{2^n-1}h_i}.
$$

In all our experiments we use normalized iris images and the corresponding normalized occlusion masks calculated by the open-source OSIRIS software \cite{osiris}. Each normalized image has a resolution of $512\times64$ pixels, which translates to 512 sampling points along the iris circle, and 64 sampling points along the iris radius (from the pupil to the sclera). Note that iris rotation in Cartesian coordinates corresponds to a circular shift of the normalized iris image in polar coordinates. It means that even if the mutual rotation between the template and the probe is non-zero, the only thing that changes is the spatial location of elements within the normalized image, hence the resulting histograms are the same.

Rathgeb \etal \cite{rathgeb_2016} proposed to calculate histograms locally in the predefined iris image patches (however, without excluding the occluded iris areas) and to binarize the histograms to calculate a compact iris image representation. We also added this method for comparison in Section \ref{sec:results}.

\begin{figure}[!t] 
    \centering
        \includegraphics[width=0.47\textwidth]{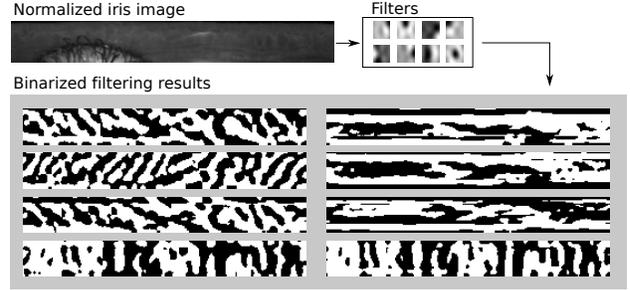}
    \caption{Iris recognition pipeline based on binary codes calculated independently for each filter. An occlusion mask (not shown here) is used in calculation of the comparison score. {\bf This iris-code-based iris matching approach is used in this paper to demonstrate effectiveness of domain-specific filters}}.
    \label{fig:iris_rec_pipeline_codes}
\end{figure}

An alternative solution, that delivers better results in our experiments, uses the binarized responses of $n$ filters directly to construct $n$ iris codes, as illustrated in Fig.~\ref{fig:iris_rec_pipeline_codes}. Each $i$-th iris code $c_i^{(t)}$ forming an iris template is compared independently with the corresponding probe iris code $c_i^{(p)}$ by calculating the fractional Hamming distance of non-occluded iris portions:

$$
\mbox{HD}_i(\theta) = \frac{\|\big(c_i^{(t)}\oplus c_i^{(p)}(\theta)\big) \cap m^{(t)} \cap m^{(p)}(\theta)\|}{\|m^{(t)} \cap m^{(p)}(\theta)\|},
$$

\noindent where $i=1,\dots,n$, $\oplus$ denotes an exclusive OR operation, $\cap$ denotes a logical AND operation, $m^{(t)}$ and $m^{(p)}$ are binary masks with 1's indicating valid areas of the template and probe samples, respectively, $\|x\|$ denotes the number of 1's in $x$, and $\theta\in\langle -11.25^\circ,11.25^\circ\rangle$ indicates the angle by which the probe image and the corresponding probe mask are rotated to find the best matching. The final comparison score may be calculated in three ways, by taking average, minimum, or maximum fractional Hamming distance out of $n$ fractional Hamming distances calculated for $n$ pairs of codes (for each $n$-th filter). This gives three another ways of calculating the comparison score considered in this work (in addition to those given by equations (\ref{zh}) and (\ref{zhn})):

\begin{equation}
z_{\mbox{\scriptsize HDmean}} = \min_{\theta}\Bigg(\frac{1}{n}\sum_{i=1}^n\mbox{HD}_i(\theta)\Bigg)
\label{zmean},
\end{equation}

\begin{equation}
z_{\mbox{\scriptsize HDmin}} = \min_{\theta}\Big(\min_{i=1,\dots,n}\big(\mbox{HD}_i(\theta)\big)\Big)
\label{zmin},
\end{equation}

\begin{equation}
z_{\mbox{\scriptsize HDmax}} = \min_{\theta}\Big(\max_{i=1,\dots,n}\big(\mbox{HD}_i(\theta)\big)\Big)
\label{zmax}.
\end{equation}

The minimum over $\theta$ in equations (\ref{zmean})--(\ref{zmax}) compensates for relative rotation of eye between two images.

\section{Related Work}
\label{sec:rw}

The present paper is related to the step of iris feature extraction, within the typical iris recognition pipeline.
Following the seminal work of Daugman~\cite{daugman_1993}, the literature of iris feature extraction was initially dominated by solutions that aimed at describing iris texture through the quantization of filter responses over iris images~\cite{kong_2010}.
While Daugman originally proposed the use of Gabor filters, other approaches inspected the employment of alternative filters, ranging from Haar wavelets~\cite{lim_2001}, wavelet packets~\cite{krichen_2004,czajka_2010}, spatial filter banks~\cite{ma_2002}, to directional filter banks~\cite{park_2003}, only to name a few.

In contrast to texture description, some researchers have lately tried to describe alternative iris features, including salient interest points~\cite{fernandez_2009, belcher_2009} and human-interpretable features~\cite{flynn_2016}.
Nonetheless, a prominent set of works still has been focusing on iris texture, specifically employing general-purpose texture descriptors such as Local Binary Patterns (LBP)~\cite{ojala_2002}, Local Phase Quantization (LPQ)~\cite{ojansivu_2008}, and BSIF~\cite{Raja_IWBF_2014, rathgeb_2016}, the latter descriptor reportedly constituting better iris recognition systems.

The idea of improving BSIF descriptor with domain-specific filters was recently verified by Ouamane et al.~\cite{ouamane_2017} in the problem of face recognition, where new filters were learned from two- and three-dimensional face images. In the particular case of iris recognition, though, the same idea is yet to be investigated, to the best of our knowledge.

Last but not least, the idea of having humans in the loop for providing human-machine collaboration towards the solution of difficult recognition problems has been applied to varied domains, such as object~\cite{branson_2010, llerena_2017, manen_2017}, face~\cite{cao_2015}, iris~\cite{mcginn_2013}, and even galaxy recognition~\cite{lintott_2008}. Human contributions may range from question-and-answer opinions~\cite{branson_2010, mcginn_2013, cao_2015}, to predefined type selection~\cite{lintott_2008}, to gaze point collection~\cite{llerena_2017}, and to free shape annotation~\cite{manen_2017}.

Relating this paper to previous work, this is (to our knowledge) first work to show that re-training BSIF filters to the iris image domain improves performance over the default filters. This is also the first work engages human subjects in constructing iris feature extractors, using both human gaze tracking and manual image annotation.

\section{Experiments and Results}

In this Section we summarize the experimental setup and report the obtained results.

\subsection{Filter Training}
\label{sec:training}

The task of filter training regards the computation of a set $F$ of filters from a training set $P$ of image patches, by maximizing the statistical independence of the responses of the filters belonging to $F$, when applied over the patches $P$.

\paragraph{Iris Patch Extraction.}
The set $P$ of image patches is obtained from $\mathcal{S}_{\mbox{\scriptsize domain}}$ set, hence making the proposed solution specific to the domain of iris recognition. Moreover, rather than randomly sampling the irises belonging to $\mathcal{S}_{\mbox{\scriptsize domain}}$, we rely upon the opinions and behavior of human subjects for finding regions of interest.

\begin{figure}[!t]
    \centering
    \frame{\includegraphics[width=0.47\textwidth]{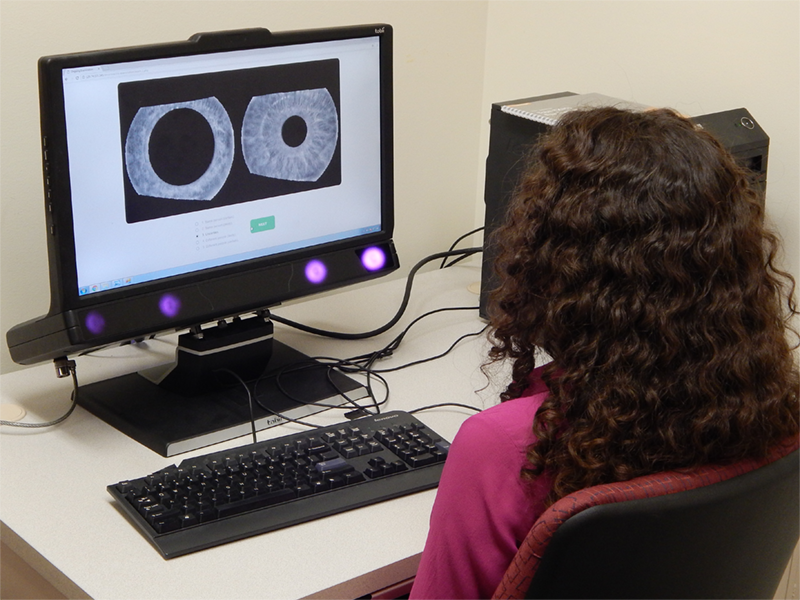}}
    \caption{Hardware and software setup for gathering people's decisions, annotations, and gaze points.
    We used a dedicated hardware for eye tracking and gaze collection (\emph{Tobii Pro TX300} eye tracker~\cite{tobii_2018}), and a specially designed web application for gathering decisions and manual annotations.}
    \label{fig:user}
\end{figure}

\begin{figure}[!t]
    \centering
    \frame{\includegraphics[width=0.47\textwidth]{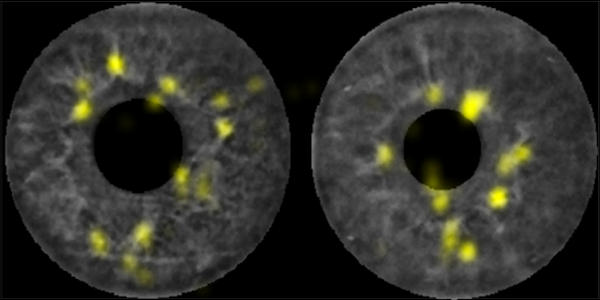}}
    \caption{A sample visualization of gaze region detection. Yellowish regions indicate areas at which a subject spent more time when performing the iris recognition tasks. These areas were then used in generation of iris image patches.}
    \label{fig:gaze}
\end{figure}

\begin{figure}[!t]
    \centering
    \frame{\includegraphics[width=0.47\textwidth]{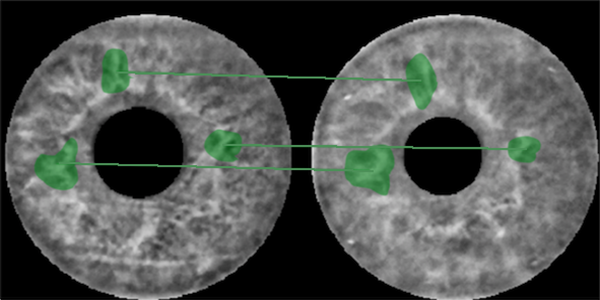}}
    \caption{Detailed screen for manual annotation.
    Users can annotate and connect arbitrary shapes over the two irises, meaning that they probably refer to the same iris region of interest (highlighted areas).}
    \label{fig:annot}
\end{figure}

For gathering people's input, we set up an experiment with 86 volunteers, comprising university staff and students who were not necessarily familiar with the problem of iris recognition.
Through a specially designed web application interface and a dedicated hardware to perform eye tracking (Fig.~\ref{fig:user}), ten pairs of iris images belonging to $\mathcal{S}_{\mbox{\scriptsize domain}}$ were presented to each volunteer, depicting either authentic or impostor cases. The $\mathcal{S}_{\mbox{\scriptsize domain}}$ set was composed of 800 iris images (400 pairs) of various quality and origin.
To increase the variation of samples, and to make the iris recognition problem more challenging to humans, $\mathcal{S}_{\mbox{\scriptsize domain}}$ was composed of a mixture of matching and non-matching pairs representing images that were ``easy'' or ``hard'' to match, according to OSIRIS software, images of the same irises but with excessive difference in pupil dilation, images of different eyes from identical twins and triplets, and even post-mortem iris images.

The task of volunteers participating in the experiment was to decide, for each pair, if the displayed irises either belonged to the same eye or not. \add{While doing this, their gaze was automatically collected through an eye tracker device, whose data was later processed using a spatio-temporal criterion to localize important regions, according to the gaze persistence over the same neighborhood (depicted in Fig.~\ref{fig:gaze}.} Next, subjects were asked to manually annotate each iris pair with arbitrary shapes, highlighting regions that supported their decisions. As one might observe through Fig.~\ref{fig:annot}, matching regions were annotated as green connections, according to the subject's desire.

Only $512 \times 512$-pixel masked clean near-infrared irises were presented to the volunteers, guaranteeing that decisions were made solely based on the iris texture, rather than on periocular information. All image segmentations in $\mathcal{S}_{\mbox{\scriptsize domain}}$ were done manually to guarantee the correctness of masking periocular regions and more challenging areas, such as irregular eyelashes, specular reflections or cornea wrinkles in post-mortem iris images.

\begin{figure}[!t]
    \centering
    \begin{subfigure}{0.3\textwidth}
        \centering
        \frame{\includegraphics[width=\textwidth]{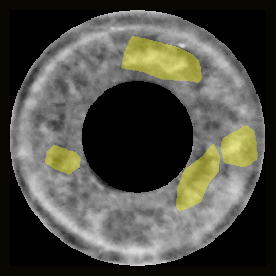}}
        \caption{}
    \end{subfigure}\\\vskip3mm
    \begin{subfigure}{0.47\textwidth}
        \centering
        \frame{\includegraphics[width=\textwidth]{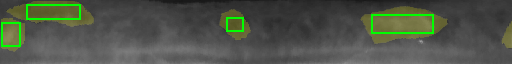}}
        \caption{}
    \end{subfigure}
    \caption{Normalization and extraction of human-annotated patches: an example of original human annotations (a), the normalized iris image and normalized annotated areas along with the rectangular representations of the human-annotated patches used in training (b).
    \add{Gazed-based regions of interest were extracted in the same way.}}
    \label{fig:patch}
\end{figure}

Prior to extracting the patches $P$, both manual annotations and gaze-based regions of interest --- which were initially established on original iris images for making them more intelligible to humans --- had to be normalized with Daugman's rubber sheet transformations~\cite{daugman_2004}. This allows to follow the typical iris-code-based recognition pipeline applied in earlier works.
After normalization, patches were extracted from each human-inspired region, according to the largest inscribed rectangle, as shown in Fig.~\ref{fig:patch}. To ensure the quality of the annotations and of the gaze-based regions of interest, patches were extracted only from {\bf genuine iris pairs} that were {\bf correctly classified by subjects}. This is important since humans are not considered to be skilled in iris recognition task, hence using only correctly classified examples increases the probability to learn valuable information from human subjects.

\paragraph{Filter Set Computation.}
Once the human-inspired iris patches $P$ are available, one can compute the set $F$ of ICA filters, which, by definiton~\cite{Kannala_ICPR_2012}, depends on two parameters: the size $l$ of the filters and the number $n$ of filters.
In this work we investigate $12$ filter sizes, namely $l \in \left\{5, 7, 9, 11, 13, 15, 17, 19, 21, 27, 33, 39\right\}$, and $n \in \left\{5, 6, 7, 8, 9, 10, 11, 12\right\}$ filters in one set. This gives 96 different sets of filters $F$ (compare this to 60 different sets available in the original BSIF).

Depending on $l$, each iris patch is randomly sampled with squares whose sides are equal to $l$. Sizes larger than the patch itself are discarded. As a consequence, we end up with 142,852 square sub-patches extracted from manual annotations, and 143,177 square sub-patches extracted from gaze-based regions of interest, which are then used for filter training. We never mix annotation-based and gaze-based patches in the experiments. The computation of filters is done with the \emph{scikit-learn Fast ICA} implementation~\cite{scikit}. The extracted human-inspired iris image patches are available along with the paper for those who want to apply other software packages or methodology for filter training.

\begin{figure*}[!htb]
    \centering
        \includegraphics[width=\textwidth]{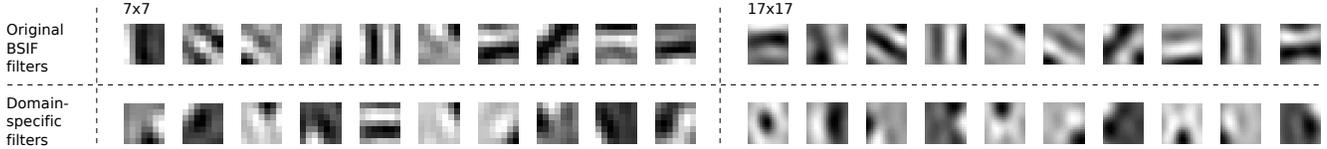}
    \caption{Example sets of ten general-purpose and domain-specific filters (trained on iris patches) at two different scales, $7\times7$ and $17\times17$ pixels. Note that standard BSIF filters are in most cases close to edge detectors, while new, domain-specific filters are close to filters sensitive to dot-like shapes. This difference can be especially observed for larger filters shown on the right.}
    \label{fig:filters}
\end{figure*}

Fig.~\ref{fig:filters} compares two sets of $n=8$ original BSIF and our new filters, for two example scales ($7\times7$ and $17\times17$). One interesting observation is that original BSIF filters are in most cases similar to edge detectors at different angles (top row in Fig.~\ref{fig:filters}). However, new filters suggest more sensitivity to dot-like features, which are more common in iris texture than straight edges.

\subsection{Filter Selection and Matching Strategy}
\label{sec:selection}
We use a separate set of 1,812 iris images $\mathcal{S}_{\mbox{\scriptsize train}}$ representing 453 different irises of 330 new subjects (compared to $\mathcal{S}_{\mbox{\scriptsize design}}$ set) to select the best set of filters and the best matching strategy out of five strategies presented in Sec. \ref{sec:irisRecognition}. $\mathcal{S}_{\mbox{\scriptsize design}}$ mixes samples acquired by both AD 100 and LG 4000 iris sensors.

\begin{figure}[!htb]
    \centering
        \includegraphics[width=0.235\textwidth]{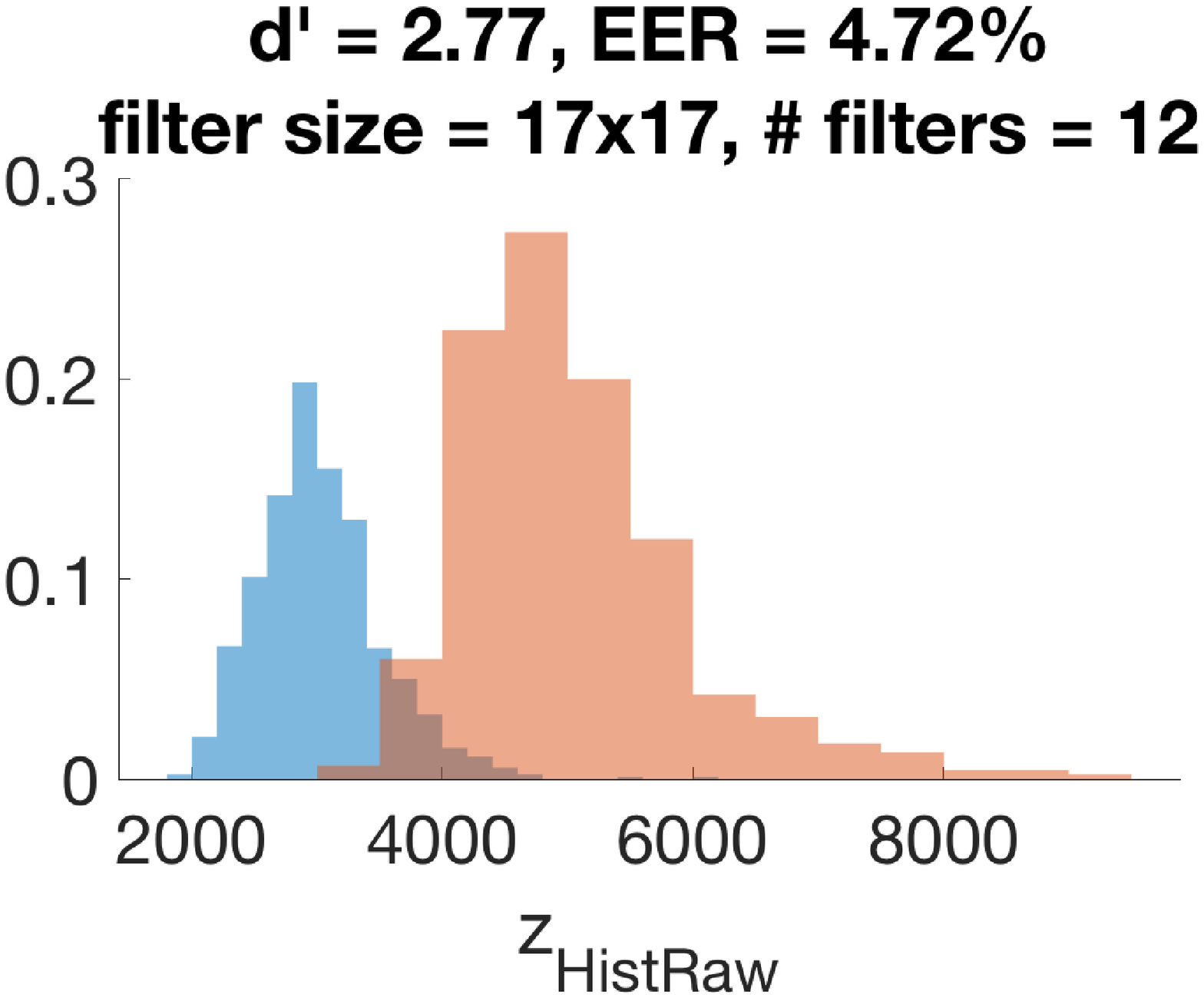}\hfill
        \includegraphics[width=0.235\textwidth]{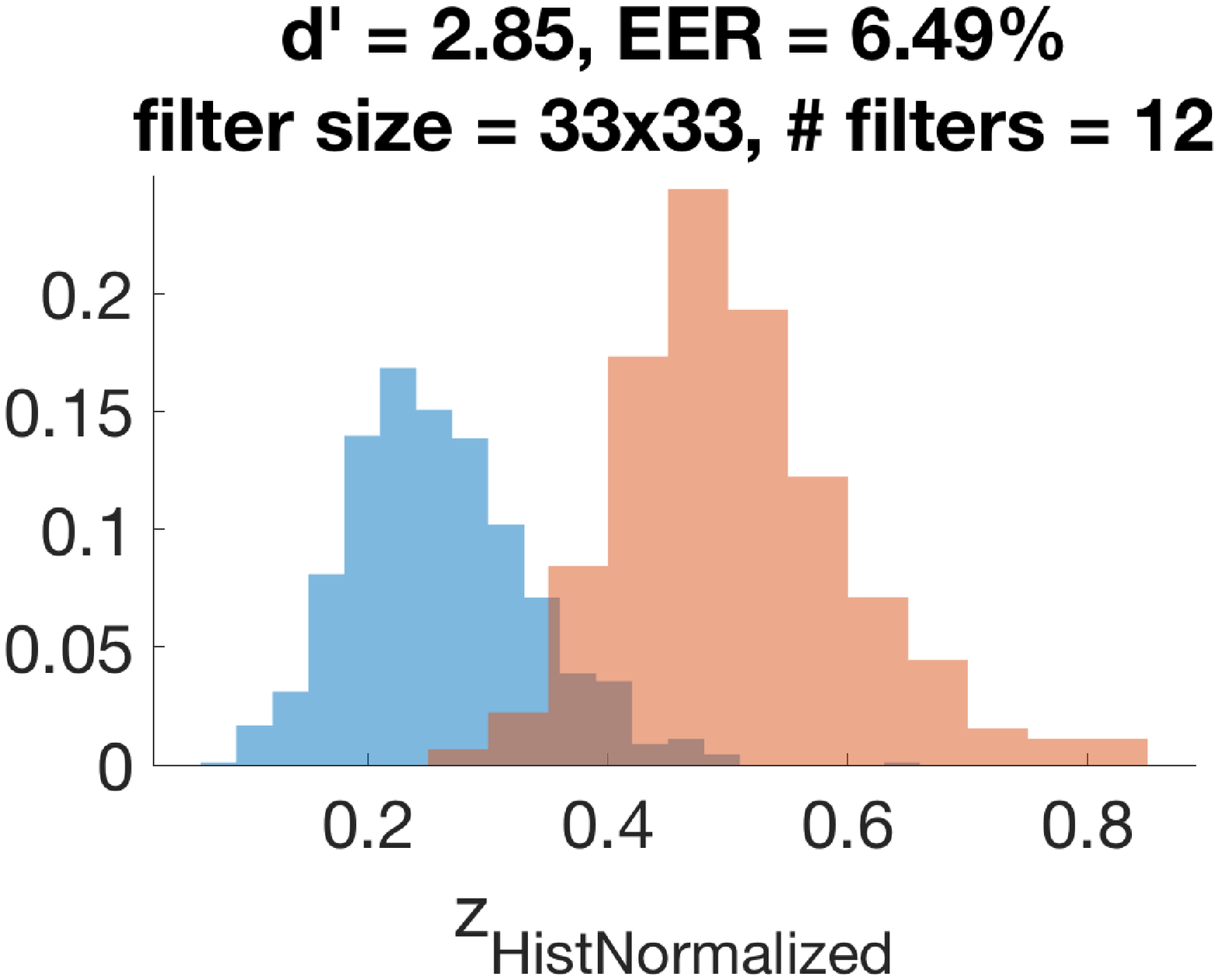}\\\vskip2mm
        \includegraphics[width=0.235\textwidth]{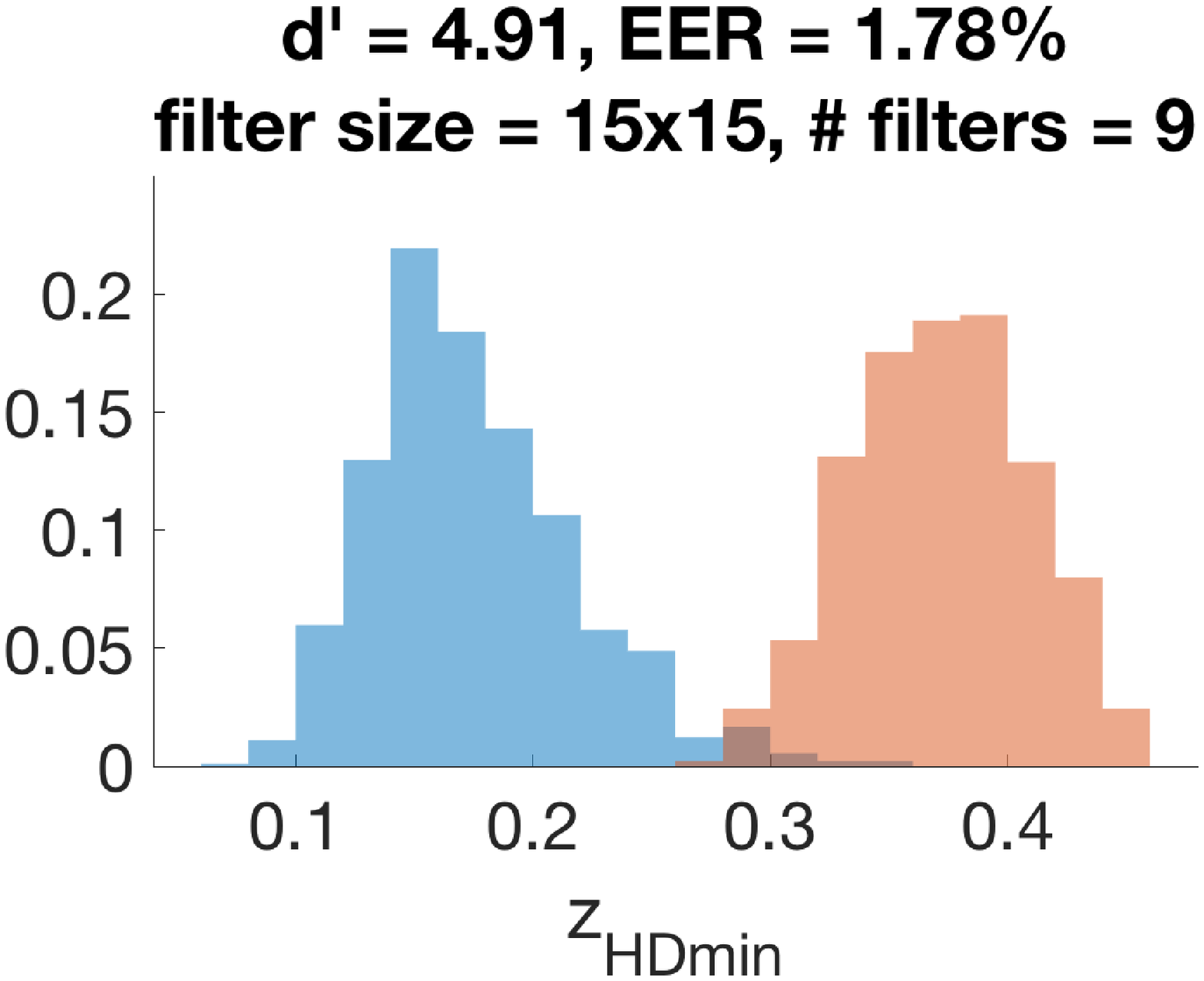}\hfill
        \includegraphics[width=0.235\textwidth]{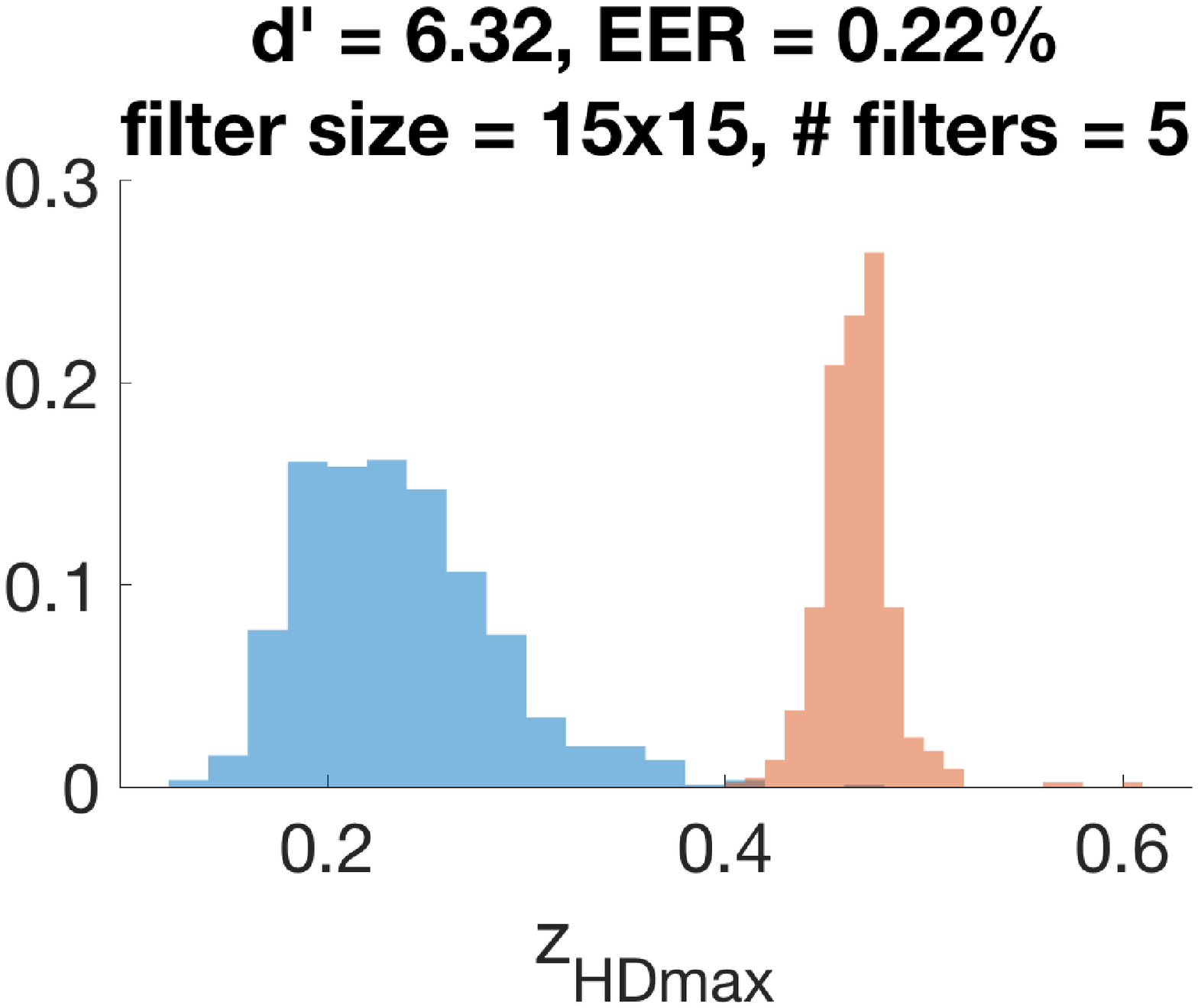}\\\vskip2mm
        \includegraphics[width=0.235\textwidth]{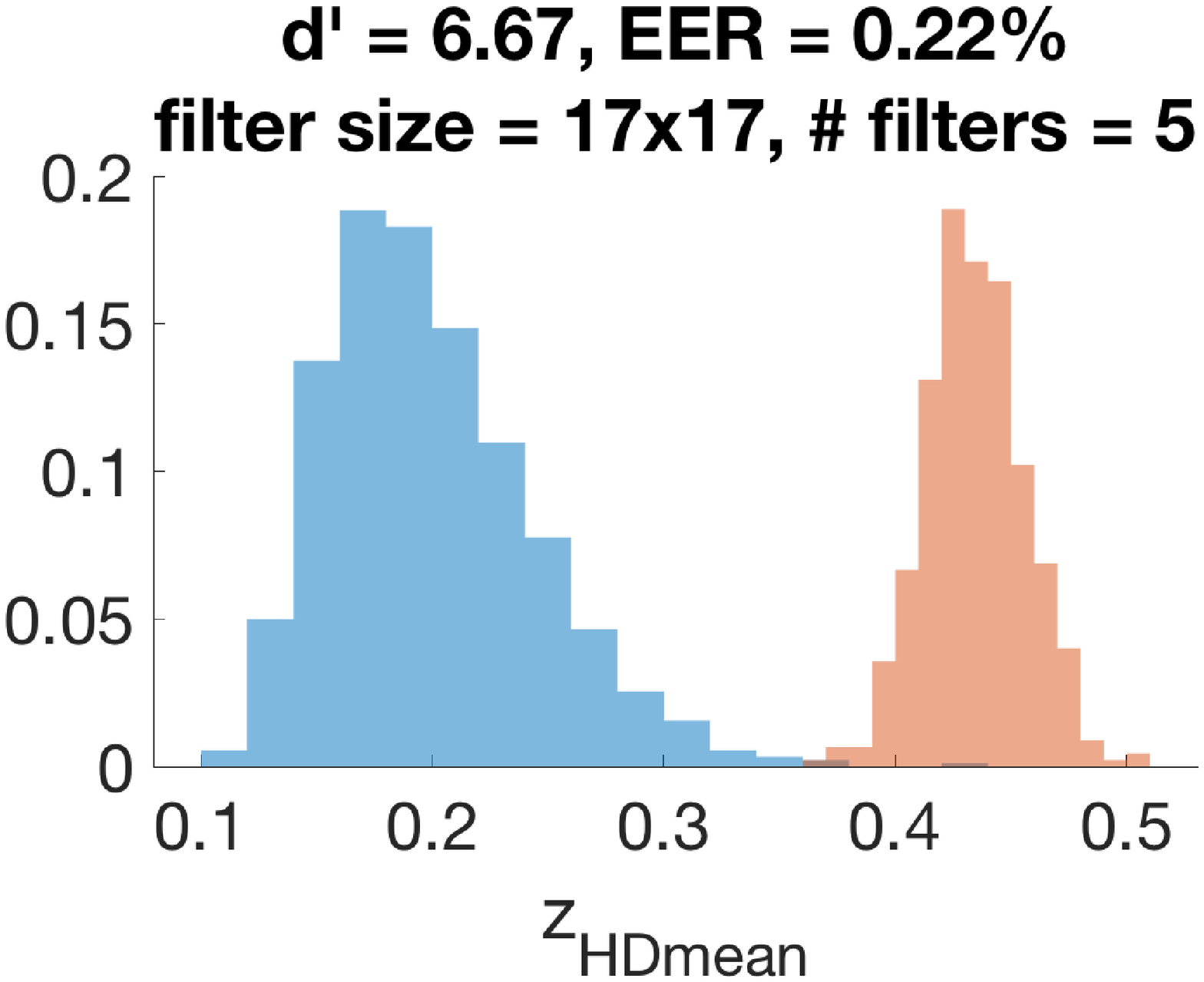}
    \caption{The best results for filters learned from eye tracker-based iris image patches for five different matching strategies defined by equations (\ref{zmean})--(\ref{zmax}), and found with samples taken from $\mathcal{S}_{\mbox{\scriptsize train}}$ dataset. Each plot informs on the best filter set, $d'$ and EER achieved on $\mathcal{S}_{\mbox{\scriptsize train}}$.}
    \label{hists}
\end{figure}

Selection of the best filter sets, \ie $n$ and $l$, was performed for standard BSIF filters and newly designed filters independently.
Hence, our search was a Cartesian product of $60\times5 = 300$ combinations for standard BSIF filters, and $96\times5 = 480$ combinations of domain-specific filters derived from a single set of image patches. We took only a single image pair for a combination of eye (left/right), subject and sensor (AD 100 / LG 4000) when generating comparison scores, in order to minimize statistical dependence among scores. This ended up with 906 genuine comparisons and 453 impostor comparisons available for each evaluation. Some approaches out of the total number of 780 possibilities achieved Equal Error Rate = 0. So, to make a further comparison among all solutions, we accumulated both sample mean and sample variances of scores in the form of the so-called {\em d-prime} (or {\em decidability}, or {\em detectability}):

\begin{equation}
d' = \frac{|\smean{\xi}_g-\smean{\xi}_i|}{\sqrt{\frac{1}{2}\big(\svar{\xi}_g+\svar{\xi}_i\big)}},
\label{eqn:DPrime}
\end{equation}

\noindent where $\smean{\xi}$ and $\svar{\xi}$ denote sample mean and sample variance of $\xi$, respectively. Explaining shortly, the further the mean values are located for same variances, the better is the separation of distributions. Similarly, keeping the same $\smean{\xi}_g$ and $\smean{\xi}_i$ and simultaneously narrowing $\svar{\xi}_g$ and $\svar{\xi}_i$ one may get higher level of distinction between genuine and impostor scores. Consequently, the value of $d'$ estimates the degree by which the distributions of $\xi_g$ and $\xi_i$ overlap. For uncorrelated random variables, $d'=\sqrt{2}\zeta$ where $\zeta$ is a standardized difference between the expected values of two random variables, often used in hypothesis testing. 

Figure \ref{hists} presents distributions of genuine and impostor scores, along with the best combinations of filters, for newly designed filters for eye tracker-based image patches and for all five matching strategies. These plots suggest that a solution calculating mean fractional Hamming distance between binary codes works better (\ie achieves higher $d'$) than solutions based on histograms. \add{This matching strategy wins when standard BSIF filters or filters learned from iris data are used}, and thus only this approach is used in final evaluations on the $\mathcal{S}_{\mbox{\scriptsize test}}$ set.

\subsection{Testing and Comparison of Solutions}
\label{sec:results}

\begin{figure*}[!htb]
    \centering
    \begin{subfigure}{0.245\textwidth}  
    \centering
        \includegraphics[width=\textwidth]{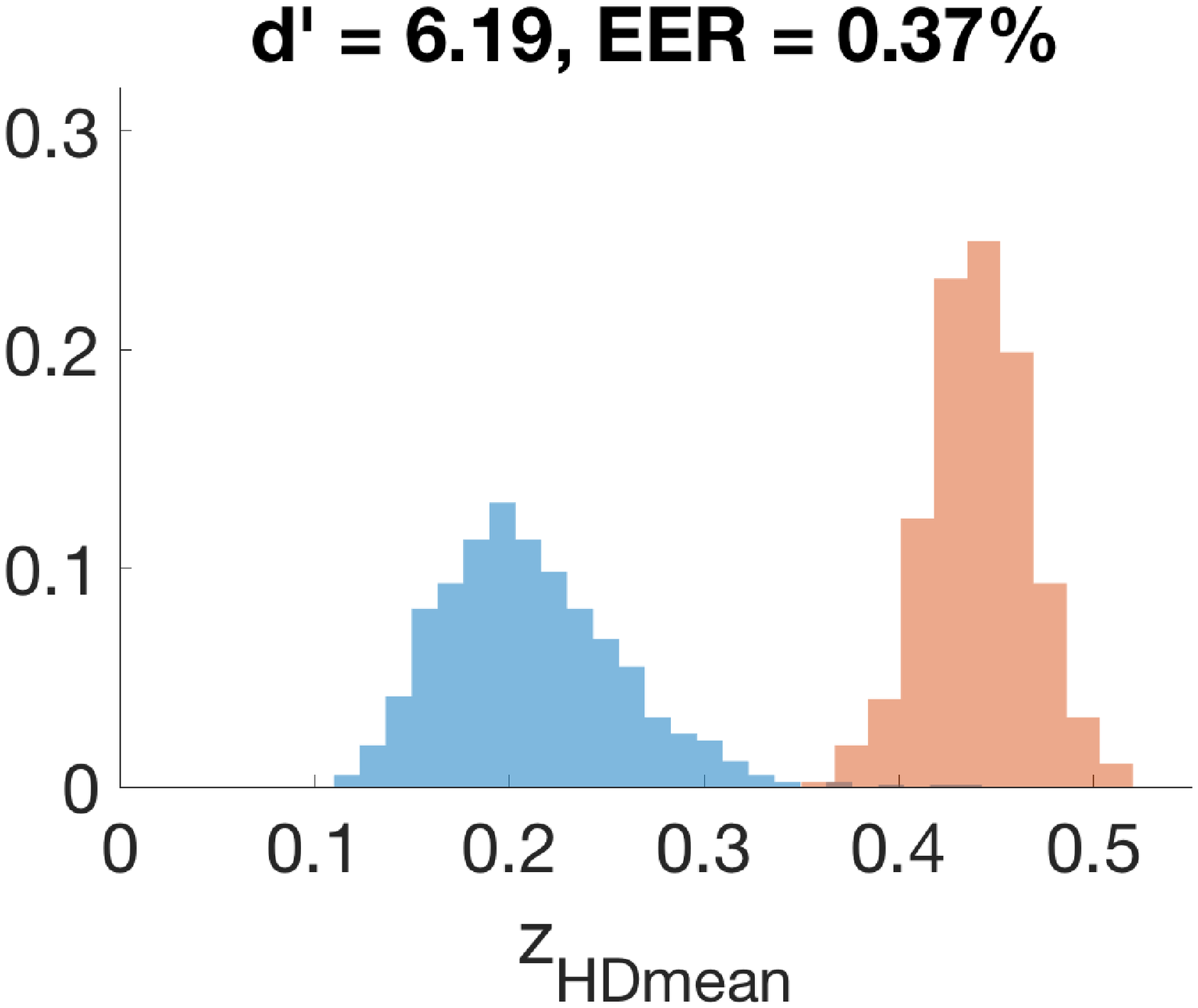}
        \caption{}
    \end{subfigure} 
    \begin{subfigure}{0.245\textwidth}
    \centering
        \includegraphics[width=\textwidth]{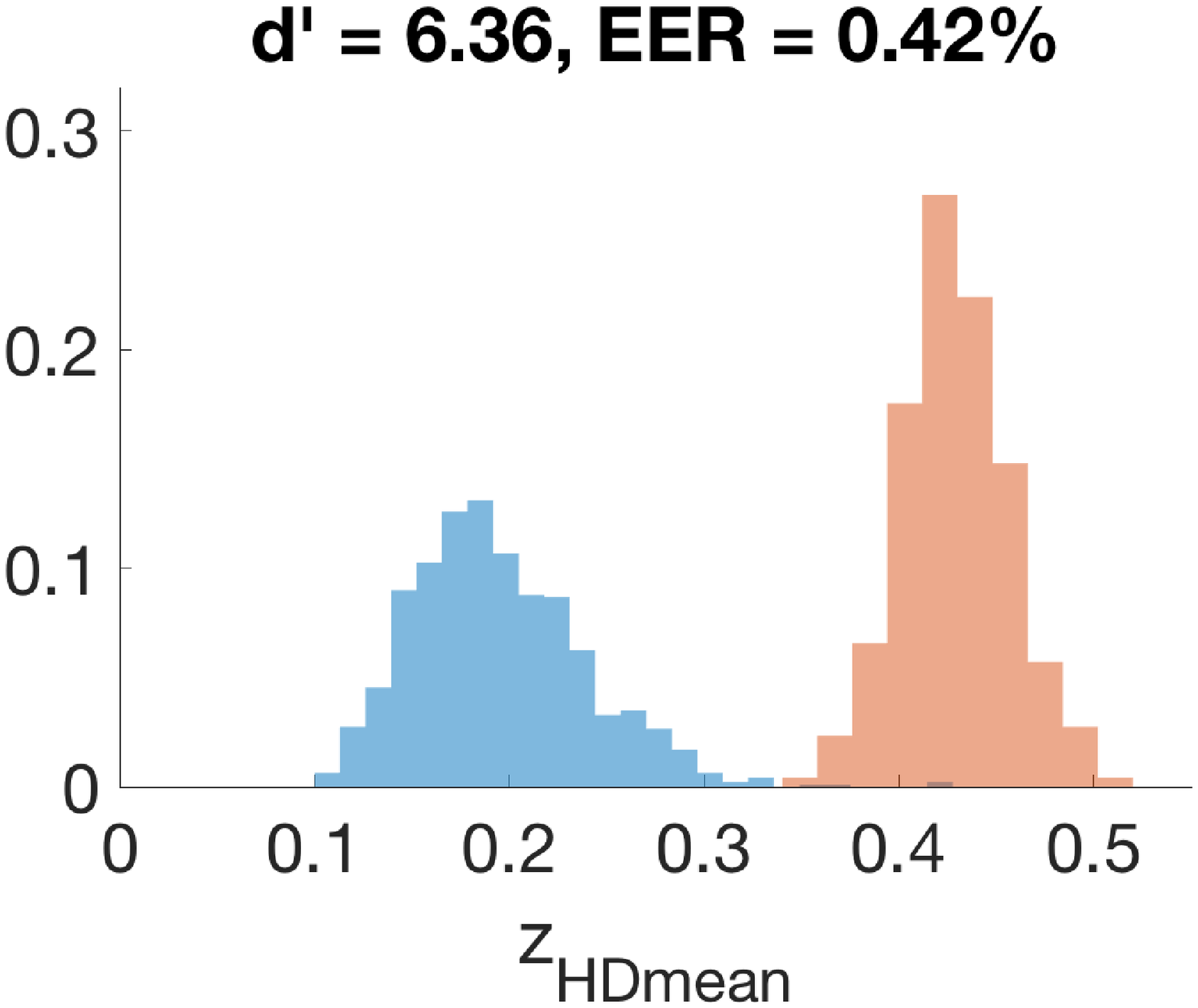}
        \caption{}
    \end{subfigure}
    \begin{subfigure}{0.245\textwidth}
    \centering
        \includegraphics[width=\textwidth]{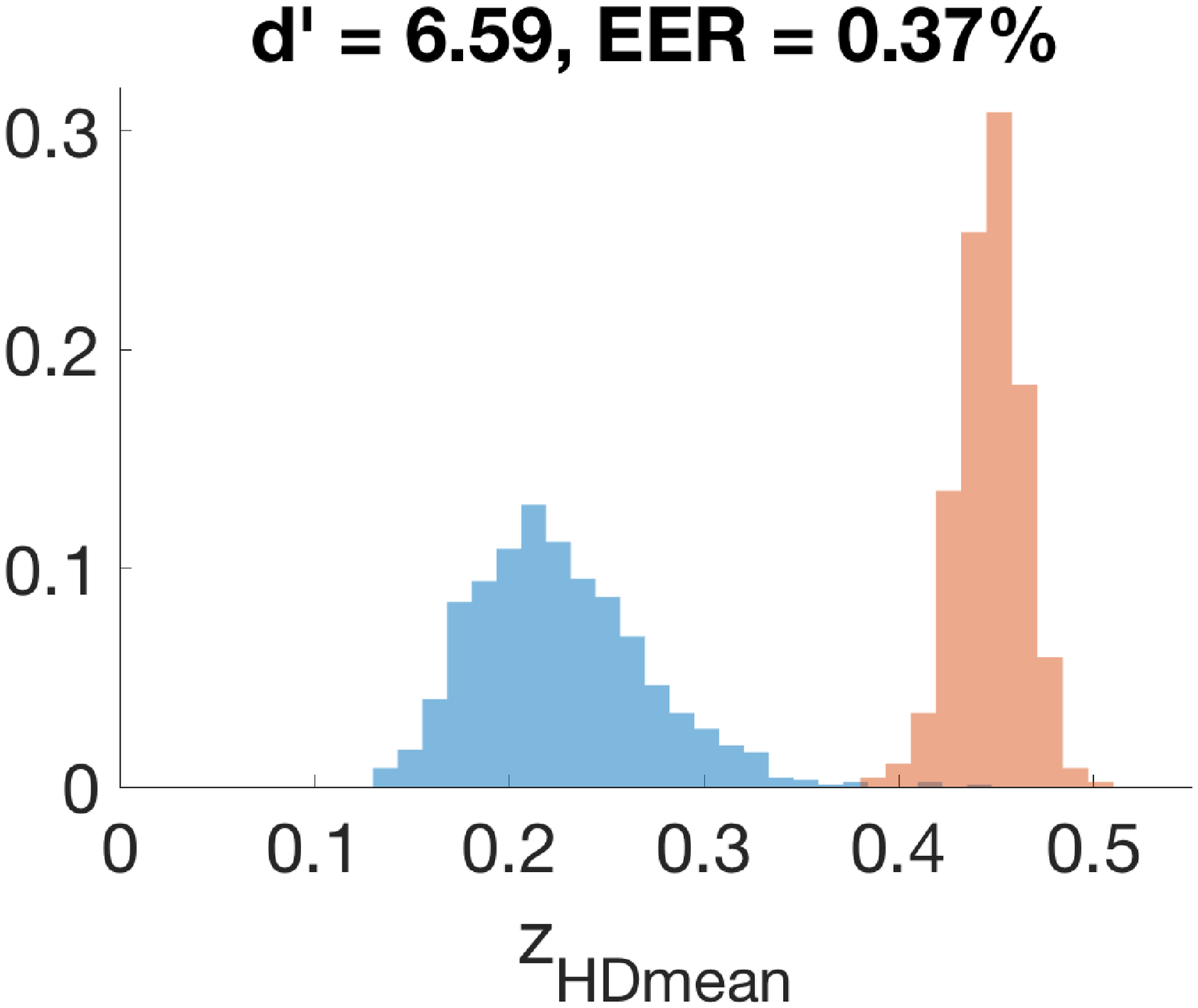}
        \caption{}
    \end{subfigure}
    \begin{subfigure}{0.245\textwidth}  
    \centering
        \includegraphics[width=\textwidth]{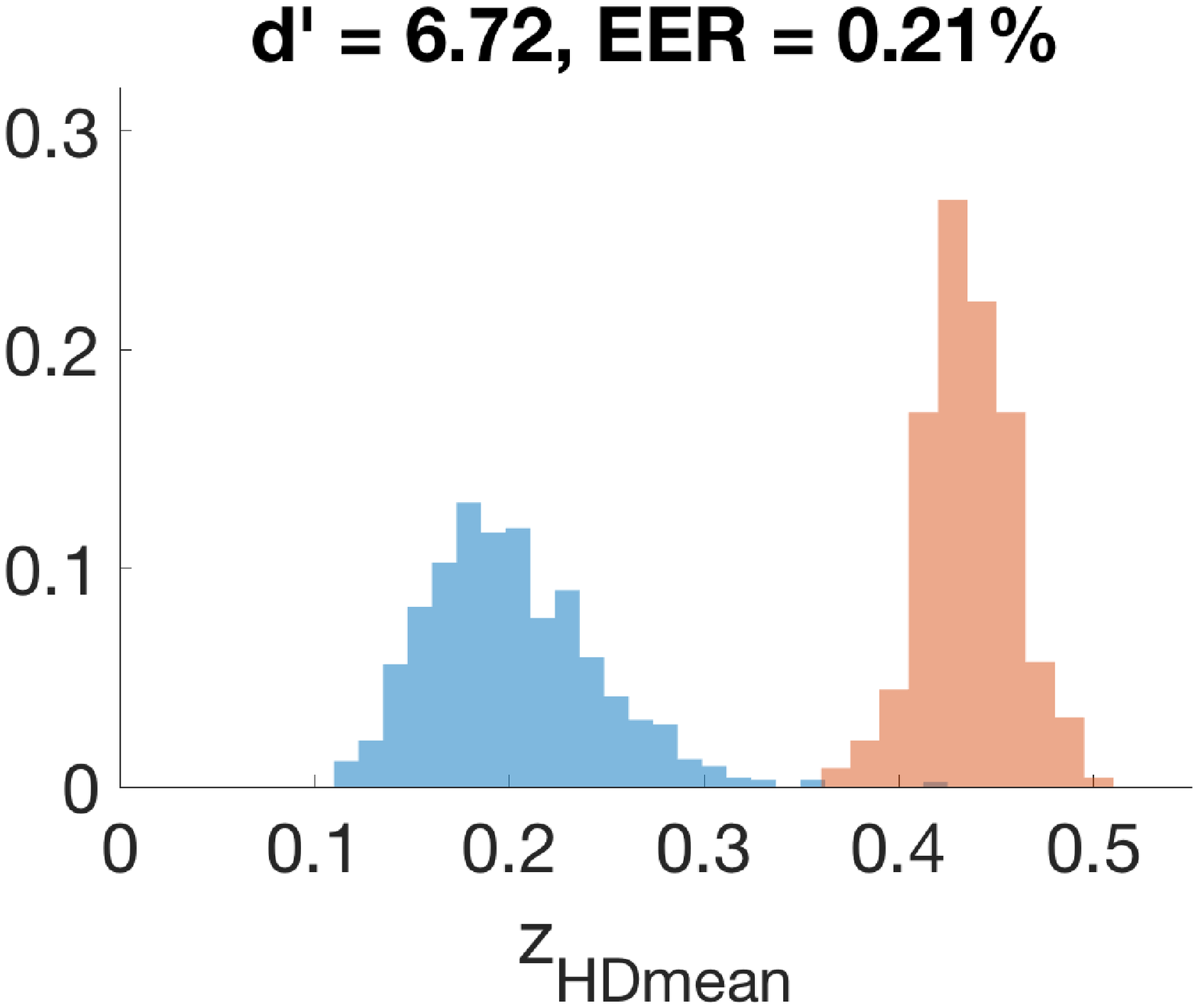}
        \caption{}
    \end{subfigure}
    \caption{Genuine and impostor score distributions on the test set $\mathcal{S}_{\mbox{\scriptsize test}}$ for all four BSIF-based methods considered in this paper: a) standard BSIF filters, b) filters learned from random iris image patches, c) filters learned from human-annotated iris areas, d) filters learned from eye tracker-based iris patches. EER and d' are also presented for each method.}
    \label{fig:s3AllHistBSIF}
\end{figure*}

\begin{figure}
    \begin{subfigure}{0.235\textwidth}
    \centering
        \includegraphics[width=\textwidth]{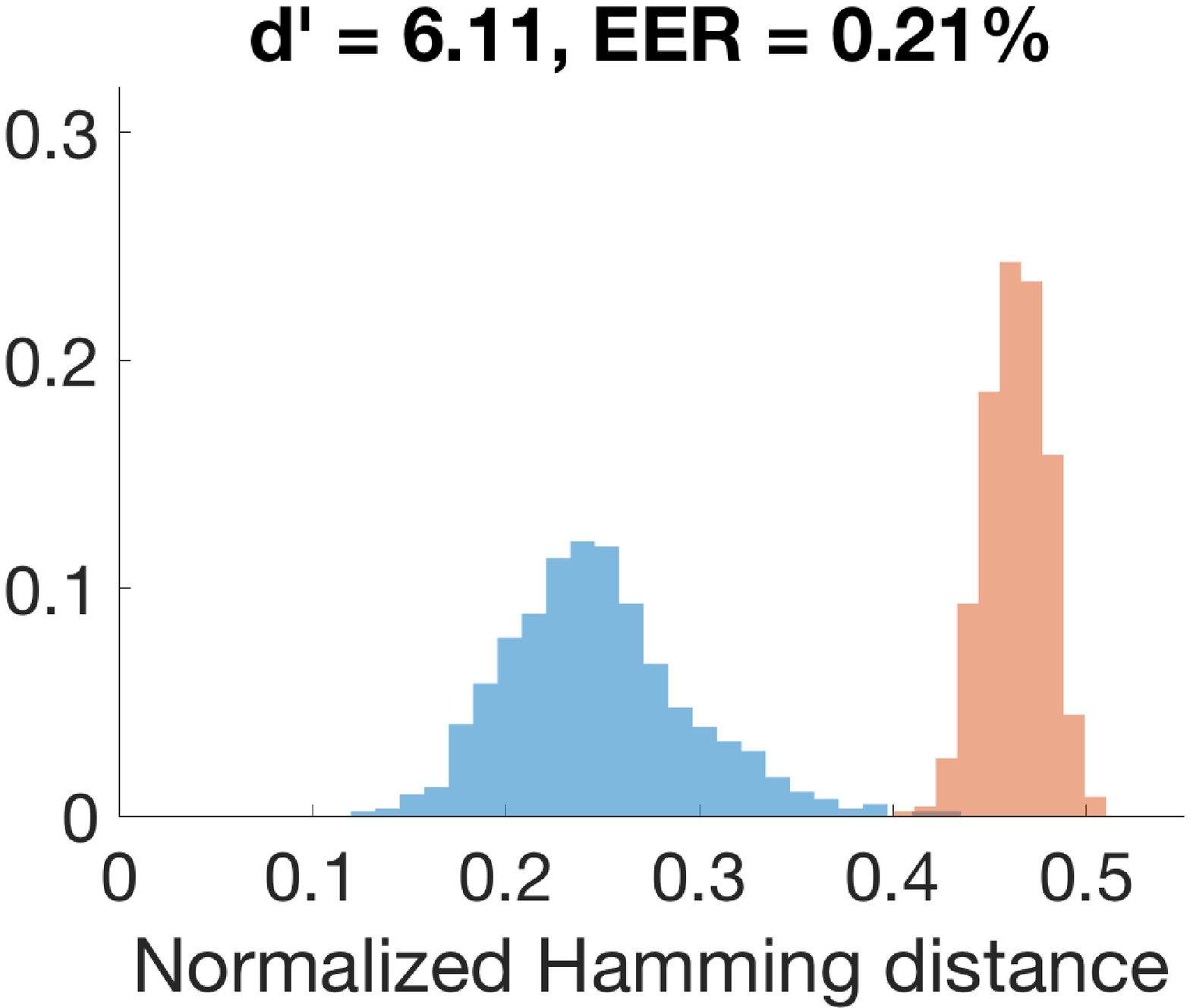}
        \caption{}
    \end{subfigure}
    \begin{subfigure}{0.235\textwidth}
    \centering
        \includegraphics[width=\textwidth]{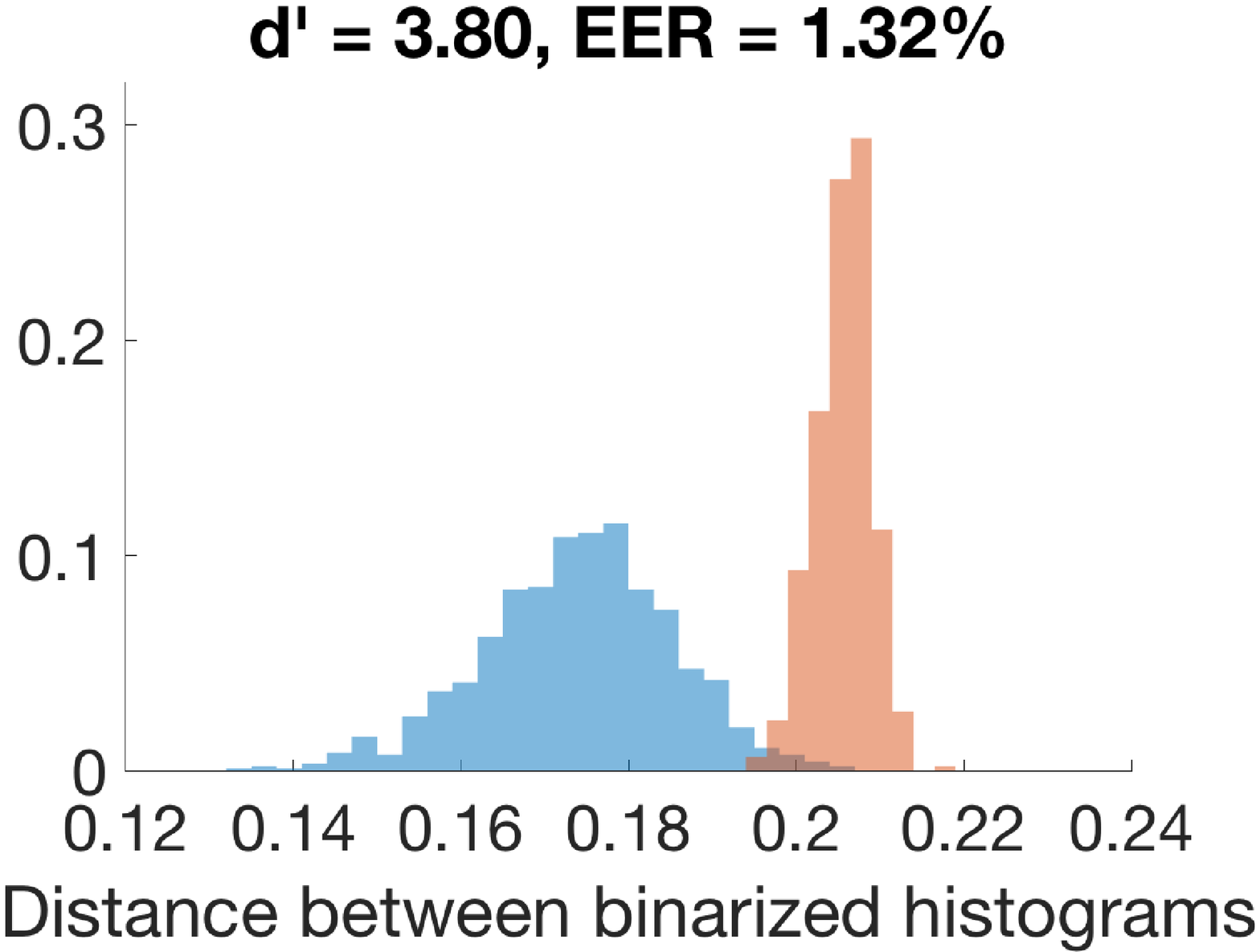}
        \caption{}
    \end{subfigure}
    \caption{Genuine and impostor score distributions on the test set $\mathcal{S}_{\mbox{\scriptsize test}}$ for two third-party methods: a) OSIRIS \cite{osiris} and b) USIT \cite{USIT2}.}
    \label{fig:s3AllHistOther}
\end{figure}

To verify the hypothesis about the advantage of domain-specific filters, we conduct final experiments on an additional set $\mathcal{S}_{\mbox{\scriptsize test}}$ of 1,900 iris images acquired from the next 330 subjects (different than the subjects represented in $\mathcal{S}_{\mbox{\scriptsize train}}$). Figures \ref{fig:s3AllHistBSIF} present comparison score distributions, EER and d' obtained on the test set for all BSIF-based methods. To relate these approaches to other methods, we add results obtained for the OSIRIS matcher~\cite{osiris} and Rathgeb~\etal~\cite{rathgeb_2016} approach distributed along with the USIT package \cite{USIT2}, as shown in Fig.~\ref{fig:s3AllHistOther}.

The first observation is that domain-specific filters based on random patches allow to achieve better $d'$ than standard BSIF filters (cf. Fig. \ref{fig:s3AllHistBSIF}a-b). The second observation is that learning new filters from patches derived from human-annotated regions results in further increase of $d'$ (Fig. \ref{fig:s3AllHistBSIF}c). Finally, one can notice the highest $d'$ and the lowest EER obtained for filters learned from eye tracker-based iris image patches (Fig. \ref{fig:s3AllHistBSIF}d). The former method also outperforms OSIRIS and USIT algorithms. This suggests that using image areas that the subjects look at when performing the iris recognition task in filter design allows to obtain the most discriminative iris features.

\begin{figure}[!htb]
    \centering
        \includegraphics[width=0.47\textwidth]{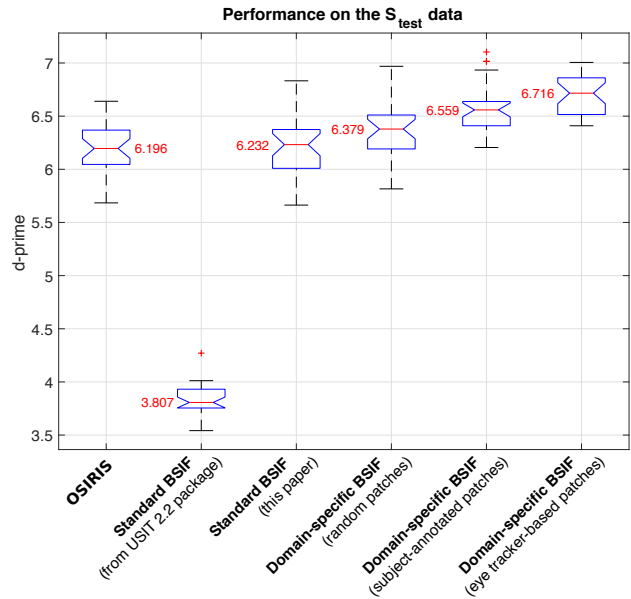}
        \caption{Boxplots illustrating distributions of $d'$ obtained in 30 independent experiments on $\mathcal{S}_{\mbox{\scriptsize test}}$ dataset for four different iris recognition methods. Median values are shown in red, height of each boxes corresponds to an inter-quartile range (IQR) spanning from the first (Q1) to the third (Q3) quartile, whiskers span from Q1-1.5*IQR to Q3+1.5*IQR, and outliers are shown as red crosses. Notches represent 95\% confidence intervals of the median. Note the highest $d'$ (\ie the best performace) for newly designed, domain-specific filters.}
    \label{fig:allMethods}
\end{figure}

To verify if the observed differences are statistically significant we randomly selected (with replacement) 30 sets of genuine and impostor comparison scores calculated for $\mathcal{S}_{\mbox{\scriptsize test}}$ samples. We decided to evaluate each subset of scores by analyzing $d'$ since the evaluated methods performed perfectly (EER=0) on some of the generated 30 sets of comparisons. The resulting boxplots gathering all 30 random selections of scores for all methods considered in the paper are presented in Fig.~\ref{fig:allMethods}. One may note superiority of domain-specific filters. We applied one-way ANOVA\footnote{One of the ANOVA requirements is normality of tested distributions. However, Szapiro-Wilk test for normality suggests that there is no reason to reject a hypothesis on normality of all distributions of $d'$ used in this evaluation ($0.25 < p\mbox{-value} < 0.7$, depending on the approach).} to verify if differences in the performance observed in Fig.~\ref{fig:allMethods} are statistically significant. Obtained $p\mbox{-value}<0.04$ for all pairs of approaches listed along the horizontal axis in Fig.~\ref{fig:allMethods} indicates that observed differences in $d'$ are statistically significant (at the assumed significance level $\alpha=0.05$).

A possible explanation of a significantly lower performance of the method included into the USIT package, as observed in Fig.~\ref{fig:allMethods}, may be related to the fact that this method does not incorporate occlusion mask. All other methods evaluated in this paper exclude non-iris portions of the iris annulus from feature comparison. 

Hence, the answer to the first question posed in the introduction is affirmative: {\bf the adaptation of BSIF filters to an iris recognition domain does allow to extract more discriminative iris image features than standard BSIF filters.} Also the answer to the second question is affirmative: {\bf a careful selection of iris image patches, based on regions used by humans performing iris recognition task, does allow to increase the iris recognition performance when compared to using filters trained on randomly selected iris images patches.}

\section{Conclusions}
\label{sec:conclusions}

Works in the literature have consistently verified the effectiveness of BSIF texture descriptors in iris recognition. The following question has arisen: if a set of filters learned from arbitrary images is useful for iris recognition, what improvements could result if such filters were replaced by iris-sourced ones? This paper shows, through a comprehensive three-step subject-disjoint experimental setup, that learning new iris-specific BSIF filters, especially employing iris image regions important to humans, results in a statistically significant improvement in iris recognition accuracy. To our best knowledge, this is the first time humans were put in the loop for learning domain-specific BSIF filters. For completeness, we also presented results for filters trained on iris patches selected randomly.

As an important contribution of this work, we make our iris-sourced BSIF filters and all iris images patches used in training available to the community, along with the database of test images. We also offer the source codes of the iris recognition method based on newly designed filters that proved the best in our evaluations. \add{These allow for a full reproduction of the results presented in this paper.}

The next interesting research step would be applying these domain-specific filters in other areas related to iris recognition, for instance presentation attack detection, in the algorithms that already used standard BSIF successfully. Since we keep the format of the re-trained filters identical as for standard filters, this replacement is trivial.

\section*{Acknowledgements}
The authors would like to cordially thank Dr.~Sidney D'Mello and Mr.~Robert Bixler for making their laboratory and eye tracker device available for conducting the experiments.

{\small
\bibliographystyle{ieee}
\bibliography{ref}
}

\end{document}